**Automated Analysis, Reporting, and Archiving for Robotic Nondestructive Assay of Holdup Deposits – 19508**

Heather Jones *, Siri Maley *, Kenji Yonekawa *, Mohammadreza Mousaei *, J. David Yesso, David Kohanbash *, William Whittaker *
* Carnegie Mellon University

**ABSTRACT**
To decommission deactivated gaseous diffusion enrichment facilities, miles of contaminated pipe must be measured. The current method requires thousands of manual measurements, repeated manual data transcription, and months of manual analysis. The Pipe Crawling Activity Measurement System (PCAMS), developed by Carnegie Mellon University and in commissioning for use at the DOE Portsmouth Gaseous Diffusion Enrichment Facility, uses a robot to measure Uranium-235 from inside pipes and automatically log the data. Radiation measurements, as well as imagery, geometric modeling, and precise measurement positioning data are digitally transferred to the PCAMS server. On the server, data can be automatically processed in minutes and summarized for analyst review. Measurement reports are auto-generated with the push of a button. A database specially-configured to hold heterogeneous data such as spectra, images, and robot trajectories serves as archive.

This paper outlines the features and design of the PCAMS Post-Processing Software, currently in commissioning for use at the Portsmouth Gaseous Diffusion Enrichment Facility. The analysis process, the analyst interface to the system, and the content of auto-generated reports are each described. Example pipe-interior geometric surface models, illustration of how key report features apply in operational runs, and user feedback are discussed.

**INTRODUCTION**
Miles of contaminated pipe remain as legacy of gaseous diffusion enrichment of uranium. As enrichment facilities are decommissioned, every foot of this pipe must be measured, each measurement must be tied to a precise segment location, and each segment must be tracked. In the past, this has taken hour after hour of manual labor at elevation to measure pipes from the outside and months to analyse data and compile official reports. The Pipe Crawling Activity Measurement System (PCAMS), developed by Carnegie Mellon University and in commissioning for use at the DOE Portsmouth Gaseous Diffusion Enrichment Facility, uses its RadPiper robot (see Fig.1) to measure Uranium-235 from inside pipes and automatically log the data. The PCAMS Post-Processing Software (PPS), discussed in this paper, analyzes robot data in minutes, facilitates analyst review, and auto-generates official reports for approved robot runs. PCAMS builds from a prior proof-of-concept robot [1,2], but the post-processing work has progressed considerably since that time. While this paper focuses solely on the PPS, companion papers present other aspects of PCAMS, including a system overview [3]; details of the RadPiper robot design, function, and performance testing [4]; and technical details of the radiometric method [5].

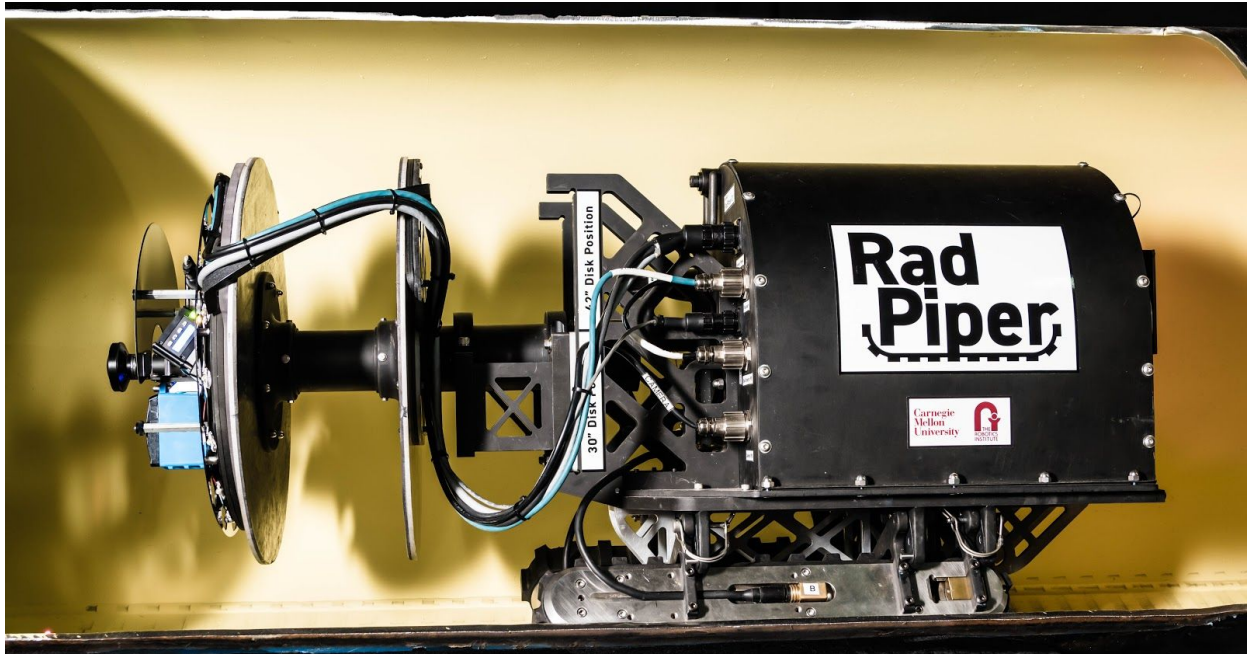

Fig.1. The PCAMS RadPiper robot

While the In Situ Object Counting System (ISOCS) [6] and Holdup Measurement System 4 (HMS4) [7] systems include several software tools for collecting measurement data and analyzing holdup measurement results, they require human participation to position instruments, and thus they also needs human input to link measurements to particular pipe segments and describe their geometry. Because PCAMS measures many pipe segments in a single run of a RadPiper robot, PCAMS PPS can automatically localize radiation measurements to pipe segments while modeling their geometry.

This paper is organized as follows. The Concept of Operations of PCAMS is described, as well as an overview of what the post-processing software does. The processing done to analyze PCAMS data is described in sections on Data Synchronization and Localization, Processing Radiometric Data and Quality Control, and Supplementary Exhibits. The Analyst Interface and Auto-Generated Reporting are each described, followed by a short section on CONDA Integration that describes how PCAMS works with the Portsmouth CONDA database. A discussion of User Feedback is followed by a section on Example Supplementary Exhibits. Conclusions, References, and Acknowledgements follow.

**CONCEPT OF OPERATIONS**
In normal pipe operations at the DOE Portsmouth site, an operator downloads a measurement request from Portsmouth's CONDA database to a PCAMS tablet. This measurement request contains one or more jobs, each with an associated building, unit, cell, pipe item identification (ID), and other information. When the operator selects a job and initiates a run of the RadPiper robot for that pipe, data from the measurement request are transferred to the robot over WiFi and logged in the data file for the run (called a "bag" file). Operator notes about the run and about the nearest column ID are also sent to the robot at this time.

The RadPiper robot performs a quality control (QC) check and a contamination check. Once the robot successfully completes its pre-run QC check, it begins measuring the selected pipe. It proceeds down the pipe at an even speed, steering to keep itself centered. The robot stops when its forward-looking safeguard sensors detect an obstacle, such as a valve, fitting, or particularly large deposit, or when it reaches an operator-specified maximum distance, whichever comes first. It then pauses for 15 seconds before reversing out of the pipe at the same even speed. After one or more runs, the operator turns off the robot and removes a USB (universal serial bus) jump drive from a protective capsule. The drive is scanned for contamination and then taken to the PCAMS Server. A technician logs into the PCAMS server and uploads bag files from the USB drive.

The PPS unpacks a bag file into a set of text CSV files and images, and it reads identifying information about the bag file, such as which robot collected it and which pipe was measured. Each run, contained within one bag file, is considered one "batch" of measurements. A unique batch number is created at this point using the robot ID and the measurement time. Bag files that have been uploaded and unpacked become available to an analyst for processing. When the analyst logs in to the PCAMS server, he or she can select from available runs. After selecting a run, the analyst inputs a few parameters, (if changes are needed from the defaults), and hits the 'Process' button. This initiates the automated analysis part of the PPS. When the analysis completes, results are displayed to the analyst in the user interface. This includes "flags" that indicate potential problems with particular segments or with the entire batch. The analyst can look at data in the results summary section of the user interface or look at details for individual segments. An analyst can also generate a draft report. The analyst must clear all flags before he or she can lock the batch. This requires providing a comment to justify why the flag can be cleared or rejecting segments for which flags cannot be cleared. Once the batch is locked, it can be reviewed by a Program Manager (PM) in the user interface. The PM can either approve the batch and push a button to get an auto-generated final report, or pass the batch back to the analyst for further review. Once the batch is final, a file containing results formatted for upload to the CONDA database can be generated.

### Auto-Analysis and Reporting

PCAMS PPS reads in measurement request information, pipe geometry data, gamma spectra, robot localization sensor data, accelerometer data, and images that have been unpacked from a bag file. PPS then auto-processes this information to calculate grams per foot of U-235, adjusted for attenuation, from the gamma spectra. PPS localizes its estimations of U-235 grams per foot using factor graph optimization to compute a time history of RadPiper's position in the inspected pipe as it collected each gamma spectrum (discussed further in a later section).

In addition to this localized U-235 calculation, PPS computes results of the QC checks that RadPiper automatically conducts before and after each pipe run. PPS also reports a history of QC checks to facilitate plotting the trend of measurements over time. The results of all these auto-analysis processes are automatically formatted and reported in PCAMS nondestructive analysis (NDA) and nuclear criticality safety (NCS) Reports and displayed in the PCAMS PPS User Interface, discussed in further detail in the Analyst Interface section. PCAMS NDA Reports and User Interface displays also include supplementary information useful for review, including visual imagery, geometric heat maps, and gamma spectra. Flags that indicate potential deviation of pipe segments from the PCAMS model (e.g. a vacuum spool in the side of a pipe) are also reported, along with useful pipe batch, robot, and detector (e.g. calibration) information. The PCAMS PPS Analyst Interface facilitates uploading, storing, processing, and reviewing, and approving PCAMS auto-analyzed and reported results, including subjective analysis actions like flag clearing.

## DATA SYNCHRONIZATION AND LOCALIZATION

All data are timestamped by the robot when collected, providing a first level of data synchronization. The precise location of the robot at each time must then be computed. PCAMS PPS uses factor graph optimization to fuse localization sensor data into the interpolated position and uncertainty of RadPiper's gamma detection field of view throughout forward and reverse traversals of pipe. This method is detailed in [4]. Also, due to the length of the robot, not all sensors are in the same place at the same time, requiring adjustment for sensor offsets.

With the exception of the farthest measurement segment (i.e., the transition from forward to reverse traversal), RadPiper is continuously moving along its pipe axis throughout each NDA run. Each instantaneous timestamped sensor reading (from any sensor) is associated with a pipe location based on the position of the robot datum at that timestamp. The time series position of the robot datum along a pipe axis is determined by PCAMS localization, which reports based on this datum. Each measurement segment is defined by when the robot datum crosses into and out of that segment. The timestamps associated with this set of localization positions are used to correlate all other radiometric and geometric sensor data to the given segment. All segments in PCAMS batches are twelve inches long (within localization tolerances) except the one or two segments farthest from the launch edge. The last segment is the length of the robot's radiometric field of view (FOV), as calibrated. The preceding segment is the remaining pipe length modulo 12 inches, if the modulus is greater than or equal to three inches. If the modulus is less than three inches, it is combined into the preceding segment, making the resulting actual second-to-last segment up to 15 inches long (exclusive). PCAMS assigns consecutive ascending numbers to pipe segments starting from the launch (entrance) edge, meaning that (1) the FOV-length segment is the highest numbered and farthest from the launch edge; and (2) the 3-to-15-inch segment is the second highest and farthest. These rules of segment division are presented in Fig.2.

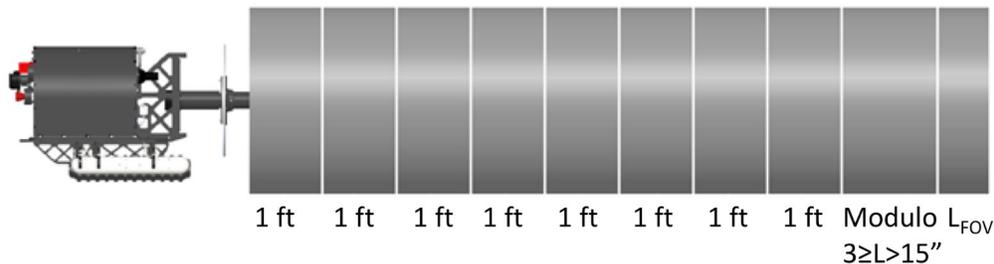

$1\ ft \quad 1\ ft \quad 1\ ft \quad 1\ ft \quad 1\ ft \quad 1\ ft \quad 1\ ft \quad 1\ ft \quad$ Modulo $L_{FOV}$
$3 \geq L > 15''$

Fig.2. PCAMS method for division of pipe batch into segments.

RadPiper's gamma detector logs a continuous accumulation of counts per spectrum channel, which RadPiper polls and timestamps at nominally 10 hertz. These timestamps are used to associate every accumulated spectrum with the localization of the robot datum. During analysis after each batch, PCAMS PPS calculates incremental spectra for time (and thus distance) window by subtracting the start spectrum from the end spectrum of the window. This method enables measurements to be precisely correlated with pipe segments, even when the location of each individual segment is not known *a priori*.

**PROCESSING RADIOMETRIC DATA AND QUALITY CONTROL**
Once gamma spectra have been associated with robot positions along a pipe, a moving sum over distance is applied. This is required because the 1/10-second spectra do not contain enough counts for useful analysis. The system then computes the counts in the 186 keV peak and associated uncertainty for each spectrum in the moving sum, resulting in a smooth curve over distance. Counts in peak numbers along the smooth curve are then converted to mass using calibrated constants, and a self-attenuation correction is applied. For each segment, the maximum of the smooth curve mass is selected for reporting. This is a conservative choice that avoids underreporting of peaks. The radiometric method is explained in more detail in [5].

**Quality Control**

PCAMS PPS includes a number of built-in quality control measures. This includes the pre- and post-run quality control (QC) and contamination checks done on the RadPiper robot and reported on the robot's tablet interface [3,4]. Due to the higher verification and validation standards required for the PPS, these checks are recomputed as part of the post-processing, and the post-processed values are taken as truth if they disagree with the values computed on the robot. There are also a number of quality control checks that only happen in the post-processing: per-segment QC check, full pipe spectrum check, and replicate check.

PPS flags the entire batch if data from RadPiper's pre- and/or post-run QC checks fail when checked against bounds for the peak width (full width at half max), peak location, or peak efficiency (gross counts) for the 60 keV peak from robot's Am-241 check source. These checks also verify that the total number of peaks detected in a spectrum is high enough (full width at half max cannot be computed if there was no detected peak to measure). PCAMS contamination checking compares the U-235 186 keV peak in the pre- and post-run QC spectra to determine if RadPiper has acquired or deposited any contamination in process piping. A contamination check failure invalidates the batch.

PPS runs the same QC check conducted pre- and post-run on the forward and reverse traversals of each segment, with the exception of allowing higher peak efficiency, which could occur due to the presence of a strong source in the pipe raising the effective background in the 60 keV region of interest used for QC checks. This is done to ensure that there was no detector failure that happened during the run which was not detected on the pre- and post-run QC checks.

PCAMS batch or 'full pipe' spectrum checking examines the accumulated spectrum from an entire pipe run for peak drift and and defocusing. In the case that the signal at the 186 keV peak is below the lower limit of detection, the full width at half max and the channel location for this peak are compared against bounds. A full pipe spectrum failure may be indicative of detector malfunction.

The purpose of the replicate check is to ensure that PCAMS measurements are sufficiently repeatable. In the standard Quality System for Nondestructive Assay (QSNDA), there must be one replicate measurement for every 20 measurements [8]. Since PCAMS measures more than 20 segments in quick succession for long pipes, the replicate strategy is somewhat different. The RadPiper robot comes out the same end of a pipe it goes in, so it measures each segment twice: once on its forward traversal and once on reverse. PCAMS replicate check compares the forward and reverse runs in two ways, "Total" and "Max". The total measured content of U-235 grams in the pipe is compared between the forward and reverse traversals, and they must match within 25% relative percent difference or 2 sigma random uncertainty. The max segment is identified in the forward traverse, and this is compared to the same segment on the reverse traverse, and they must match within 25% relative percent difference or 2 sigma random uncertainty. Failure of either part of the replicate check invalidates the run. If the replicate check passes, data from the forward and reverse runs are then averaged to get the reported U-235 mass value for each segment.

**SUPPLEMENTARY EXHIBITS**

In addition to radiometric data, PCAMS PPS provides analysts with images and geometric surface models. This section describes how images are used and how surface models are created from geometric profile data.

**Imaging**

The RadPiper robot carries a fisheye camera with LEDs (light-emitting diodes) for illumination to provide visual imaging inside pipes. These images can be used to investigate a flagged segment or to see what caused the robot to reverse (see Fig.3). During post-processing, localization data are combined with an offset between the robot position and the region viewed by the camera to assign each image a distance value.

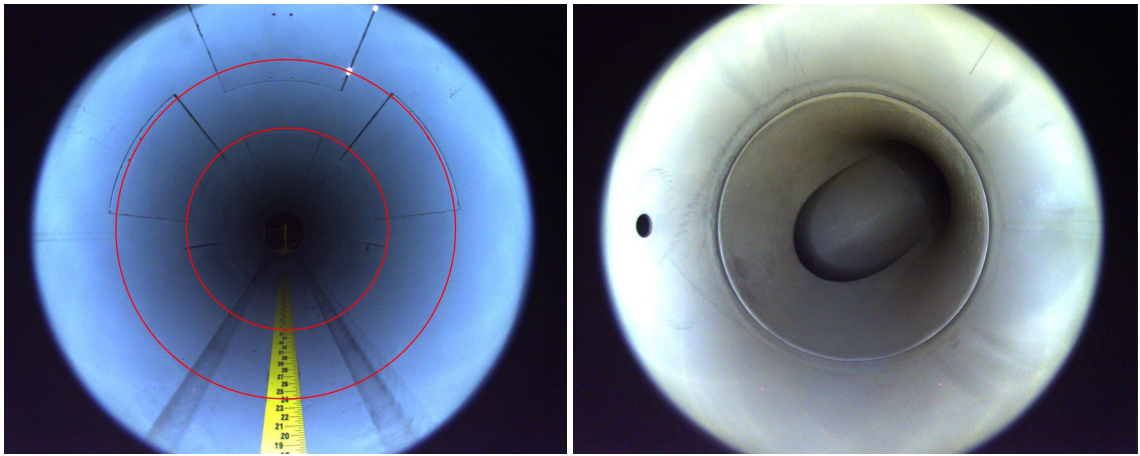

Fig.3. Left: A RadPiper image in a 30" test pipe, showing the segment of interest outlined in red. Right: A swept tee pipe fitting as viewed from RadPiper's camera

**Surface Modeling**

PCAMS PPS generates an accurate 3D model of the pipe's surface. LiDAR is used to solve this problem. Laser readings are more accurate if their angle of incidence is perpendicular to the surface. Considering near-90 degree angle of incidence at each point in the along-pipe direction, a rotating planar LiDAR (RPLIDAR) was selected as the best choice for constructing an accurate map from the environment. The RPLIDAR accuracy was also calibrated through calculating its profile over different distances and using it as a lookup table. Since the RadPiper robot platform is a transformer with variable height, RPLIDAR could not be placed at the center at all times, hence, a compromise places it such that it is 9 cm above the center for a 30-inch configuration and 7 cm below the center for a 42-inch configuration. The sensor outputs polar points in its own local frame. Given the offset of the sensor to the center of the robot, $(r_0, \theta_0)$, and a set of ranges and angles in the RPLIDAR frame, $(r_i, \theta_i)$, the algorithm first converts the raw sensor readings to the robot's frame using: $r_i = \sqrt{r_i^2 + r_0^2 + 2r_i r_0 cos(\theta_0 - \theta_i)}$ and $\theta_i = \frac{cos^{-1}(r_i cos(\theta_i) + r_0 cos(\theta_0))}{r_i}$ . The centered sensor data is then saved as a CSV file for use in geometric flagging. Thereafter, the 3D sensor data is unwrapped at $\theta = \pi$ (with $\theta = 0$ representing the bottom of a pipe) to a 2.5D heatmap. Finally, a surface is generated from those points using delaunay triangulation. The generated surface is then divided by pipe segment, tied to the associated localization results, and saved as images. Fig.4 demonstrates our test results in our test pipes where different pieces of wood are placed equidistant from each other. Tests show that the generated surface models have a fidelity of 1 cm with 2 sigma confidence.

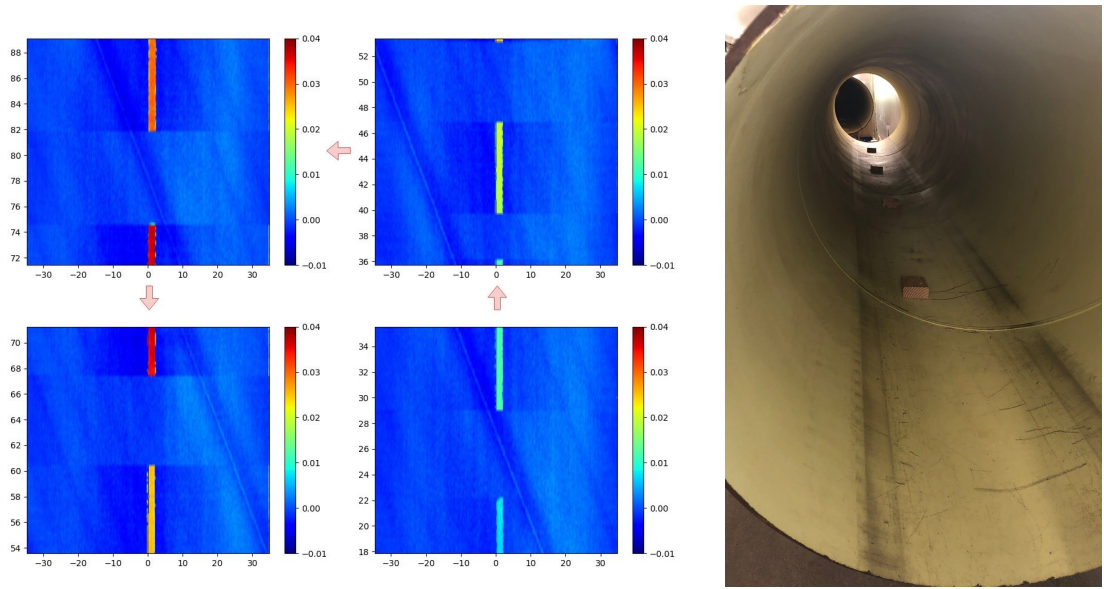

Fig.4. Left: Results of our surface modeling Right: Image of wooden test cases in test pipe

## ANALYST INTERFACE

The analyst interface allows users to process data with different parameters and interact with data. From the main screen (Fig.5), a user can modify the parameters to be used in the automated analysis. The user can also add and/or edit comments related to the pipe and pipe segments. For instance, the user may leave a note about why he/she changed the parameter for the automated analysis, or leave a comment on a pipe segment about why some flags were raised.

The analyst interface also provides some visualization in addition to the textual data shown on the main screen. The detail screen (Fig.6) shows an image, a surface model, a spectrum, and a smooth curve of mass per distance over that segment. The spectrum displayed is the one used to compute the reported mass U-235 value for that segment (generally from the forward run). One surface model, spectrum, and plot of mass per distance are represented for each segment. This is because specialists make decisions based on the per-segment information. Meanwhile, the interface shows multiple images for each segment since RadPiper typically captures multiple images within each segment. The screen allows users to interact with this information using the slider in the bottom of the screen. This allows them to smoothly go through the data while viewing information related to decision making.

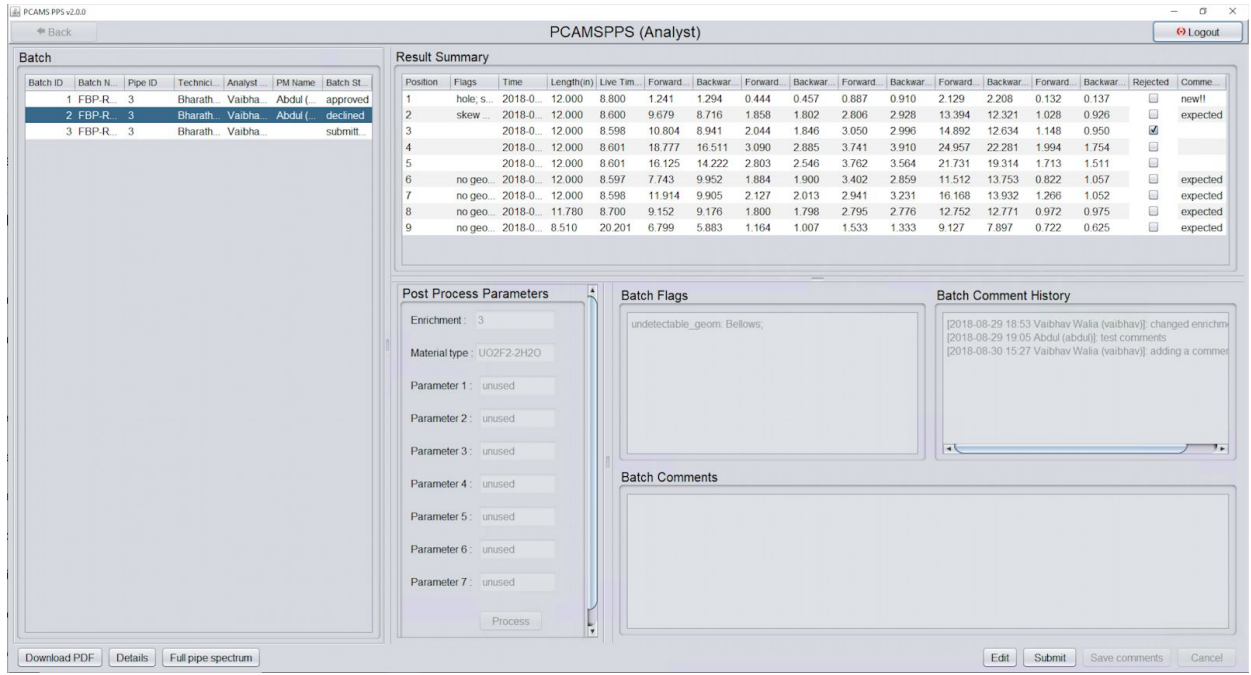

Fig.5. A main screen of the PCAMS PPS User Interface that facilitates processing and reviewing PPS analyses and reports.

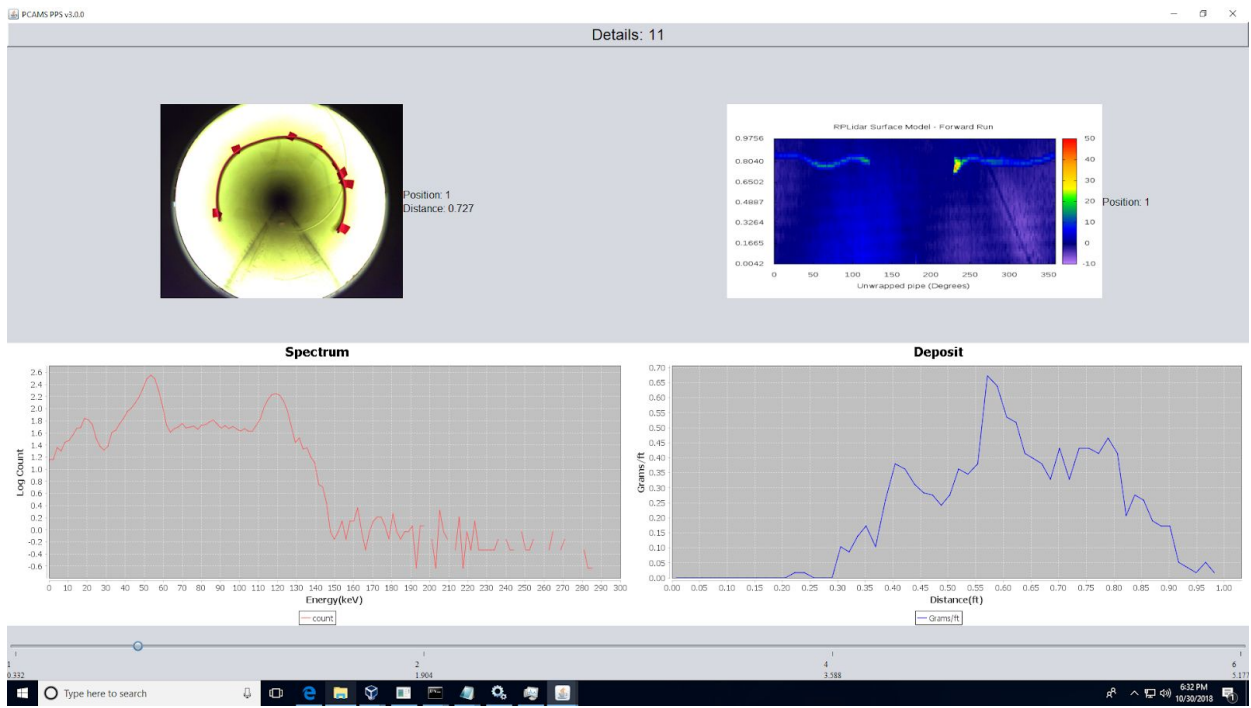

Fig.6. Details screen of analyst interface. It shows image, surface model, spectrum, and grams/foot information. The user can move the slider on the bottom of the screen to view through different sections of pipe. The screen allows users to interact with detail information about pipe segments.

**AUTO-GENERATED REPORTING**

PCAMS auto-generates NDA measurement reports. These reports include per-segment radiometric data as well as quality control trends and supplementary data used to make decisions on flagged segments.

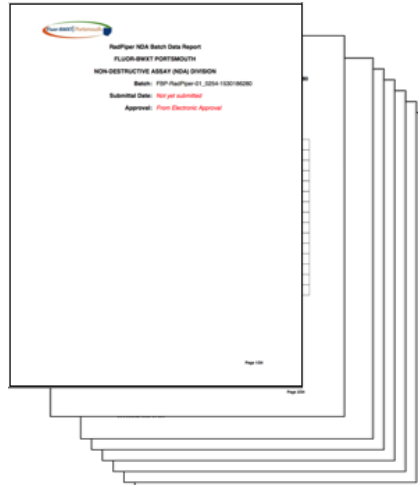

Fig.7. An auto-generated report from the PCAMS PPS, showing the cover page

A report begins with several pages of front matter, including a cover page (Fig.8), a table of contents, a table describing operational parameters such as driving speed and distance measured, and a Batch Data Report table that includes information such as file names and dates for calibration runs (see Fig.9). There is also a technical review page that evaluates whether several checks were acceptable, whether the calibration is current, and whether any flags were thrown. The central data in the report is given in one or more NDA Measurement Data Report pages that list U-235 mass, total measurement uncertainty, minimum detectable amount, and other computed parameters for each segment (see Fig.9). In the event that a segment has been rejected in the analyst interface, the text "REJECTED" will appear in place of numerical values for that segment. This is followed by pages for analyst comments on individual segments and on the batch as a whole. A replicate check page gives results of the "Total" and "Max" replicate checks. Two pages for QC check results give the trend of past QC check peak efficiency measurements and show spectra and results for the pre- and post-run QC checks for the current run.

Fig.9. Two pages of a sample PCAMS NDA Report: (left) the Batch Data Report and (right) the NDA Measurement Data Report.

The report concludes with supplemental information. This includes a raw data table with counts in peak numbers for the forward and reverse run for each segment. There is a plot as seen in Fig.10 which shows the localized mass per distance results of RadPiper's forward and reverse traversals and the per-segment report (based on the combined estimate) along with the per segment 2-sigma bounds. There are also 1-2 information pages for each flagged segment. These show an image and a spectrum plot, and, if geometric modeling is available for that segment, a surface model.

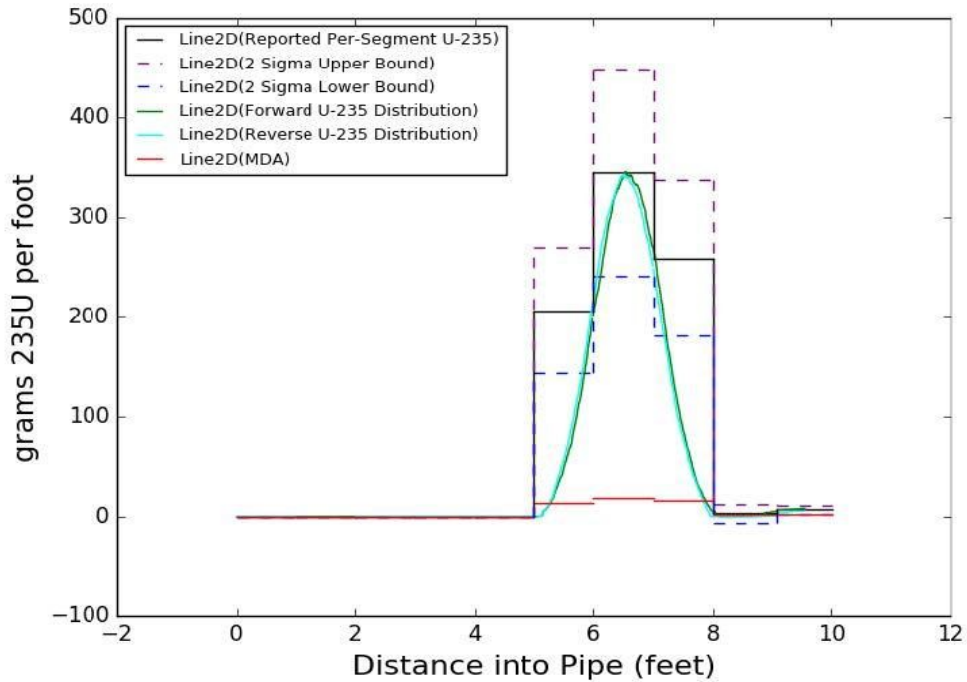

Fig.10. Plot of mass per length of pipe over 10 feet (3 meters).

In addition to the NDA report, PCAMS PPS also creates a PDF table for NCS purposes. This table lists every segment in the NDA Measurement Data Report and whether it is below the criticality incredible threshold.

**CONDA INTEGRATION**

The CONDA database is a database that stores all the pipe information within the Portsmouth facility and PCAMS interacts with it in two ways: downloading for measurement input and uploading of measured data.

Users download pipe information from the CONDA database and use it to create a measurement request. The measurement request is downloaded to the tablet which RadPiper operator checks and run the RadPiper through the pipe accordingly. In this step, pipe information such as the diameter, location, and expected length is obtained. The measurement request data are stored in the RadPiper bag file and extracted by PPS. This information can then be included in auto-generated reports.

After data has been processed and accepted, PCAMS sends the measured results to the CONDA database so that the approved information is stored and managed centrally. After the report has been approved, the approver can download the interface file from the PCAMS PPS Analyst Interface, and this file can be uploaded to the CONDA database using an upload tool. This process in an intermediate step and work currently in progress is making this process automatic by directly uploading measured results to CONDA through a dedicated API.

**USER FEEDBACK**

PCAMS PPS was fully implemented and taken through verification and validation (V&V). The system was installed on a computer that was delivered to the Portsmouth site. Through experience with the deployed software and through interaction during the July 2018 "hot test" of the RadPiper robot and other demonstrations, analysts and technicians who have used the system have provided many valuable comments. In this section, some of those comments are shared, along with how they were resolved.

**Report Approval**

Through the demonstration of automated analysis and analyst interface, the users suggested that approval of analysis should also take place in the PCAMS. This suggestion was made to follow the current commission process where a specialist checks and reports the data and another person (program manager) checks and approves or declines the report. In order to support this need, another user role was provided in PCAMS PPS with distinct privileges (analyst processes and submits the report, program manager checks and approves it). The approver and the time is captured in the interface and written in the report so that the act of signing off the report is done automatically.

**Segment Rejection**

Since the current measurement methodology cannot analyze the U-235 load for some segments with distinct features (e.g. expansion joints, different sized pipes), users suggested having the ability to reject some segments within a pipe. The rejected segments are ignored during analysis and data for these segments are not reported. The ability to mark segments of pipe as rejected in the interface and the ability to process them accordingly in the automated analysis were added. A batch with rejected segments can still be approved but the rejected segments need to be handled by some other method, such as manual measurement.

**Showing Whole Pipe Information**

This was a comment on the detail interface. The detail screen provides per segment information to analyze each segment in detail; however, there was no screen suitable to understand the overview of the entire pipe. The user suggested that the interface should provide the whole pipe information and so a new screen which provides the whole pipe information was introduced.

**Link and Show Calibration Information**

Users suggested that the interface should show the related calibration information. Each detector is mandated to have calibration once in a while and it is essential to store the calibration result for traceability purposes. This is presented in the report but not in the interface. Changes to support this need have not yet been incorporated; however, there is an intent to utilize the same database used by PCAMS PPS to track batches to also track calibration data in a future release. This will enable linked information to be shown on the interface.

**EXAMPLE SUPPLEMENTARY EXHIBITS**

PCAMS PPS has been used to analyze commissioning runs in test pipes with known-value encapsulated sources to evaluate how well radioactive material can be localized and measured. Fig.11 shows an image and the associated surface model from one of these runs. A vial encapsulated uranium source appears in the segment shown. Vial sources are analogs for lumps of deposit and are flagged as unmeasurable due to high self-attenuation. A void is created in the top of the test pipe by removing a cut-out panel in the segment after the one shown. The start of this void can be seen at the top corners of the surface model.

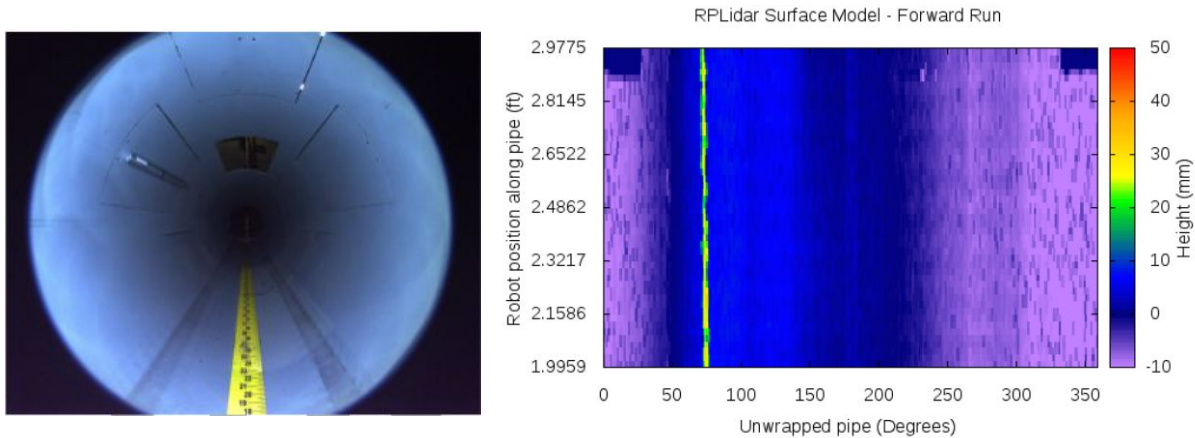

Fig.11. Image (left) and surface model (right) from a test pipe showing a vial encapsulated source.

Fig.12 shows a plot of mass per distance for one of these runs. Note that the data from the forward (green) and reverse (cyan) runs line up well.

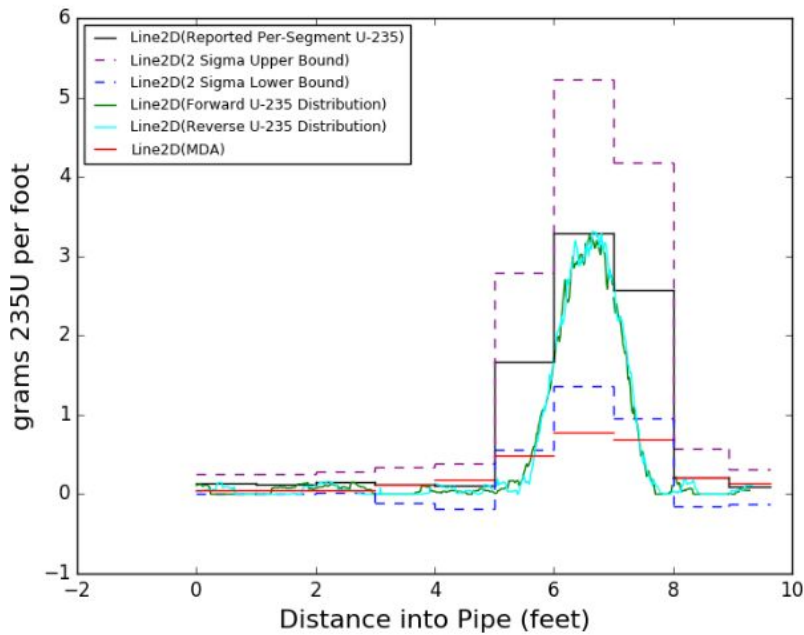

Fig.12. Plot of mass per distance from a PCAMS PPS auto-generated report for a run in a test pipe with a 3g encapsulated "hybrid tacky mat" source located from 6 ft to 7 ft, processed using hybrid tacky mat material properties

PCAMS PPS was also used to analyze data from the "hot test" of the RadPiper robot in July 2018 in which the robot measured real process piping in the Portsmouth facility. Fig.13 and Fig.14 show example images and surface models from these tests.

<Figure Redacted Pending Approval> <Figure Redacted Pending FOUO FOUO Approval>

Fig. 13. Image (left) and surface model (right) for a vacuum port in a 30" process pipe

<Figure Redacted Pending Approval> <Figure Redacted Pending FOUO FOUO Approval>

Fig.14. Image (left) and surface model (right) for a pipe joint in a 42" process pipe

**CONCLUSIONS**
A new software system for processing data from robotic pipe NDA has been designed, implemented, integrated, verified, and validated. It has been installed at the DOE Portsmouth facility and used by NDA analysts and technicians. Feedback from early users and associated system adaptations have been presented. Samples of supplementary exhibits, including images, surface models, and plots of NDA measurements by distance, have been shown for RadPiper runs in test pipes and process pipes.

There are many avenues of future work for the PCAMS PPS. These include integration of processing for calibration data into the analyst interface, automatic detection of expansion joints or other pipe features, database tracking of calibration and quality control data, improved localization methods, implementation of alternate analysis method options, and modularization and increased database integration for analysis software to enable faster re-processing. There would also be adaptation needed to handle robots for NDA of additional pipe sizes which are currently in development.

The Pipe Crawling Activity Measurement System accelerates evaluation of holdup deposit in uranium enrichment piping. This not only due to the speed of measurements, but also to the automation of analysis and report generation, which can accomplish in hours or minutes what once took weeks or months. This automation also avoids numerous opportunities for human error in data entry present in existing approaches. In all, PCAMS PPS promises to save significant time for the D&D effort at Portsmouth. Application of the system to the Paducah site is a near-term possibility, and it may someday be used at other sites around the world.

**REFERENCES**
1. H. Jones, W. Whittaker, O. Sapunkov, T. Wilson, D. Kohanbash, S. Maley, J. Teza, E. Fang, M. McHugh, I. Holst, C. Ng, R. Riddle, "Robotic NDA of Holdup Deposits in Gaseous Diffusion Piping by Gamma Assay of Uranium-235", #18331, WM (2018).

2. H. Jones, S. Maley, T. Wilson, R. Riddle, M. Reibold, W. Whittaker, D. Kohanbash, L. Papincak, W. Whittaker, "Results of Robotic Evaluation of Uranium-235 in Gaseous Diffusion Piping Holdup Deposits", #18303, WM (2018).
3. S. Maley, H. Jones, W. Whittaker, D. Kohanbash, J. Spisak, R. Boirun, A. Zhang, W. Whittaker, "Commissioning of Robotic Inspection and Automated Analysis System for Assay of Gas Diffusion Piping", #19500, WM (2019).
4. H. Jones, S. Maley, D. Kohanbash, W. Whittaker, M. Mousaei, J. Teza, A. Zhang, N. Jog, W. Whittaker, "A Robot for Nondestructive Assay of Holdup Deposits in Gaseous Diffusion Piping", #19504, WM (2019).
5. S. Maley, H. Jones, W. Whittaker, "Novel Radiometry for In-Pipe Robotic Inspection of Holdup Deposits in Gaseous Diffusion Piping", #19503, WM (2019).
6. Venkataraman, R. et al., "Validation of in situ object counting system (ISOCS) mathematical efficiency calibration software," *Nuclear Instruments and Methods in Physics Research A*, 422 (1999).
7. Smith, S.E. et al., "Holdup Measurement System 4 (HMS4) - Automation & Improved Accuracy," Proc. 45th Annual Meeting of the INMM, Northbrook, IL, INNM (2004).
8. Portsmouth-Paducah Project Office, "Quality System for Nondestructive Assay", DOE/PPPO/03-0235&D0 (2011).

**ACKNOWLEDGEMENTS**

This work was supported by DOE-EM under cooperative agreement DE-EM0004383.